\documentclass[letterpaper, 10 pt, conference]{ieeeconf}  

\IEEEoverridecommandlockouts                              

\usepackage{graphics} 
\usepackage{epsfig} 
\usepackage{amsmath} 
\usepackage{amssymb}  
\usepackage{cite}
\usepackage{color}
\usepackage[dvipsnames]{xcolor}
\usepackage{booktabs}
\let\labelindent\relax

\usepackage{bm}
\renewcommand{\vec}[1]{\mbox{\bm{$#1$}}}
\usepackage{enumitem}

\title{\LARGE \bf When We First Met:\\ Visual-Inertial Person Localization for Co-Robot Rendezvous}

\author{Xi Sun, Xinshuo Weng and Kris Kitani$^{1}$
\thanks{$^{1}$Xi Sun, Xinshuo Weng and Kris Kitani are with Robotics Institute, Carnegie Mellon University. {\tt\small xis@andrew.cmu.edu, \{xinshuow, kkitani\}@cs.cmu.edu}.}
}

\begin{document}

\maketitle
\thispagestyle{empty}
\pagestyle{empty}

\begin{abstract}
We aim to enable robots to visually localize a target person through the aid of an additional sensing modality -- the target person's 3D inertial measurements. The need for such technology may arise when a robot is to meet a person in a crowd for the first time or when an autonomous vehicle must rendezvous with a rider amongst a crowd without knowing the appearance of the person in advance. A person's inertial information can be measured with a wearable device such as a smart-phone and can be shared selectively with an autonomous system during the rendezvous. We propose a method to learn a visual-inertial feature space in which the motion of a person in video can be easily matched to the motion measured by a wearable inertial measurement unit (IMU). The transformation of the two modalities into the joint feature space is learned through the use of a triplet loss which forces inertial motion features and video motion features generated by the same person to lie close in the joint feature space. To validate our approach, we compose a dataset of over 3,000 video segments of moving people along with wearable IMU data. We show that our method is able to localize a target person with 80.7\% accuracy averaged over testing data with various number of candidates using only 5 seconds of IMU data and video.
\end{abstract}


\section{Introduction}

Person localization for a rendezvous is crucial in real-world applications such as assistive robots \cite{Kayukawa2019, Manglik2019, Guerreiro2019} and autonomous driving \cite{Wojke2017, Bewley2016, Weng2019_3dmot, Kiran2020, Wang2018, Weng2020, Leon2019, Man2019, Luo2018, Zeng2019, Weng20202, Rangesh2019, Yurtsever2019, Weng20203, Badue2019, Weng2019_mono3d, Wang2020_gnndettrk}. Consider the scenario where an autonomous vehicle rendezvous with its user for the first time. How does the autonomous vehicle localize the user without any information about what the user looks like? In this work, we consider the possibility of using the user's inertial measurement unit (IMU) data collected by her smartphone as a unique descriptor of the user's motion, which can be then used by the autonomous vehicle to localize the user with a dashboard camera. 

Prior work on person localization often utilizes visual-visual feature matching, assuming that the target person's appearance information is known in advance. However, this assumption may not always hold as it requires a data capture process prior to the rendezvous. To deal with the situation where the target person's appearance information is not available, we must rely on other sensor that can capture target person's information in the wild. We choose to use the inertial sensor as the 3D inertial measurement that describes the user's motion and can be matched with the visual motion information collected by the dash camera for person localization. Also, the user's 3D inertial measurement can be easily obtained because modern smart wearable devices such as smart-phone and smart-watch are often equipped with an inertial sensor. Moreover, due to its low dimensionality compared to visual data, we can transmit the inertial measurement to the autonomous vehicle in real time at a low cost. 

Our approach is based on visual-inertial feature matching. Specifically, we first obtain the visual motion information from the dashboard camera by computing the optical flow \cite{Chang2013} for a fixed time window. In the meantime, we obtain the motion information in 3D space measured by the IMU for the same time window. Since directly transforming the local 3D motion measurements and the 2D motion in the camera frame into same world coordinates is difficult and requires calibration of a fixed camera, we propose to learn a feature transformer based on the LSTM \cite{Hochreiter1997} and convolutional layers that can map the motion information from visual and inertial modalities into a joint feature space. The visual and inertial features are optimized using a triplet loss \cite{schroff2015} so that the learned features of the same person lie close in the joint feature space.

\begin{figure}[t]
    \centering
    \includegraphics[width=1.0\columnwidth]{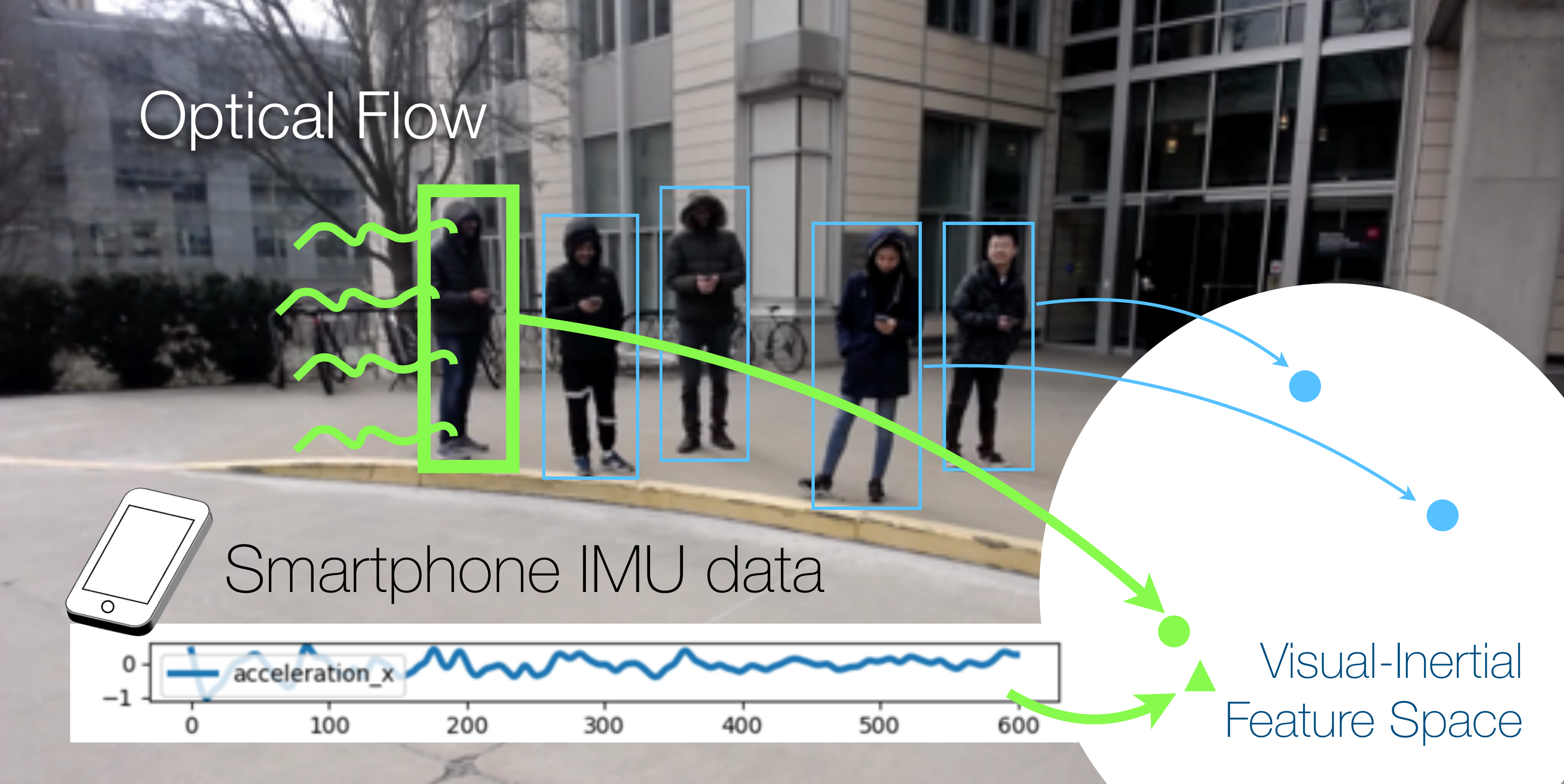}
    \vspace{-0.7cm}
    \caption{Our visual-inertial feature transformer maps IMU motion and image motion from the same person to a similar location in the feature space.}
    \label{fig:task}
    \vspace{-0.4cm}
\end{figure}

As there is no existing dataset suitable for training our feature transformer for person localization, we collect a new visual-inertial dataset containing time-synchronized video and inertial data. Our dataset has over 3,000 video segments of moving people along with their corresponding IMU data. The IMU data is collected by the smartphones held in people's hands. Different from existing visual-inertial datasets which often rigidly attach the inertial sensor on people's back \cite{Henschel2019} or body limbs \cite{Marcard2018, DIP2018, SIP2017}, we let people hold smartphones in their hands naturally to mimic the real-world scenarios. As a result, our dataset is more realistic but challenging as the location of the inertial sensor is more flexible and the motion of the inertial sensor might not always align with the motion of people's back or limbs. 

To validate our approach, we evaluate it on the test split of our visual-inertial dataset. Our experiments show that our approach is able to accurately identify a target person with $80.7\%$ accuracy on average using only $5$ seconds of IMU and video data. To summarize, our contributions are as follows: 
\begin{enumerate} \itemsep0em 
    \item \textbf{A new task, namely visual-inertial person localization}, which aims to localize the target without requiring the appearance information of the target in advance;
    \item \textbf{A new large visual-inertial dataset}, which is collected in the wild with multiple persons without fixed attachment of the inertial sensor to each person's body;
    \item \textbf{An effective approach for the proposed task}, also being the first learning-based approach for the task and outperforming competitive baselines we devised from state-of-the-art techniques.
\end{enumerate}

\section{Related Work}

\noindent\textbf{Visual-Inertial Person Localization.} To the best of our knowledge, \cite{Henschel2019} is the only work that attempted matching between visual and inertial data for person localization. First, \cite{Henschel2019} employs a visual heading network to predict person's 3D orientation with respect to the camera from a single image. Then, they match the person's 3D orientation predicted from the image with the orientation integrated from angular velocity obtained from the inertial sensor to generate image-based person predictions. To rigidly align the orientation of the inertial sensor with the person's body orientation and make the orientation prediction problem easier, \cite{Henschel2019} attaches the inertial sensor to the back of the target person. This makes \cite{Henschel2019} not applicable in the real world scenarios where the inertial sensor can be flexible. Additionally, \cite{Henschel2019} employs velocity matching between inertial and visual data to formulate trajectories of the previously generated image-based predictions. Specifically, the 3D foot position of the person is estimated from an image, which is then used to compute the 3D velocity of the target person given a pair of images. Meanwhile, the 3D velocity can be also estimated by integrating the linear acceleration from the inertial data, which can be used to match with the 3D velocity computed from the visual data. Different from \cite{Henschel2019} which employs hand-crafted inertial features (\emph{i.e.}, orientation and velocity obtained by integration) to match with the visual data, our proposed method learns to transform visual and inertial data into a joint feature space for matching. Also, our proposed method is more useful in real world scenarios as we do not restrict the placement of the inertial sensor.

\vspace{2mm}\noindent\textbf{Visual-Inertial Dataset.} Although visual-inertial person localization is under-explored in prior work, there are existing visual-inertial datasets collected for other vision tasks. The CMU Multi-Modal Activity Database \cite{DeLaTorre2008} aims to understand cooking and food preparation activities. They rigidly attach multiple IMU sensors on person's body to collect the inertial data. In the meantime, video data is also collected from multiple viewpoints. The Total Capture Dataset \cite{Trumble2017} is designed for human pose estimation. Similarly, \cite{Trumble2017} contains synchronized multi-view video and IMU data with the inertial sensor attached to the human body. However, both \cite{DeLaTorre2008} and \cite{Trumble2017} are not suitable for person localization as 1) they only collect data for one person at a time, 2) the location of the inertial sensor is fixed, and 3) the data is collected in the indoor setting. Different from existing datasets, we collect a new visual-inertial dataset with multiple persons outside and the location of the inertial sensor flexible, in order to mimic the real-world autonomous driving pick-up scenario. 

\vspace{2mm}\noindent\textbf{Visual Person Localization.} Depart from the visual-inertial person localization, prior work has investigated person localization using only visual data with the re-identification technique. The common approach is to first obtain the feature embedding from two sources of visual data (one from an unknown query person and the other from a pre-built database containing information of the target person), and then perform classification to identify if the query person is the target person. Once the target person is successfully identified, the localization is solved. To obtain effective visual embedding for identification and localization, prior work focuses on image-based \cite{Chen2019, zhou2019, Li2020} and video-based \cite{Li2019} methods for feature learning. However, visual person localization methods are only applicable when the pre-built database containing the information of the target person is available. In other words, if we do not have the target person's information in advance, we cannot solve the localization problem with only visual information but need the aid of an additional sensor. In this paper, we investigate the possibility of using the user's inertial data for localization. 

\section{Approach}

Given a video with multiple people standing or walking, and the IMU readings from a smartphone carried by a person in the scene, our goal is to identify which person in the video the IMU data belongs to as shown in Fig. \ref{fig:task}. As described above, we aim to learn a joint visual-inertial feature space in which the visual and inertial features from the same person lie close in that space.

Formally speaking, in a video segment (150 frames or 5 seconds), we denote each person in that video by an index $n \in [N]$, where $N$ is the number of people in this video segment. For each person $n$, we extract a visual feature $g_{\textrm{VIS}}$ to encode its motion in the video. Meanwhile, we extract an inertial feature $g_{\textrm{IMU}}$ of the target person from the IMU data to encode its motion in 3D space. During training, we learn a visual feature embedding function $H_{\textrm{VIS}}: g_{\textrm{VIS}} \rightarrow f$ and an inertial feature embedding function $H_{\textrm{IMU}}: g_{\textrm{IMU}} \rightarrow f$ to map both features into the same joint visual-inertial feature space. Then a triplet loss is used to force the mapped features that belong to the same person to lie close in the joint feature space. At test time, once we find the visual embedding which is the closest to the inertial embedding in the joint feature space, the target person is localized in the video.

\subsection{Visual Feature Extraction}

In order to extract people's motion feature from a video segment, we first pre-process the video by performing person detection \cite{Ren2015, Lee2016, Weng2018_r2n, Braun2019} using YOLOv3 \cite{yolov3} at all frames and then associating the detections into trajectories using a multi-object tracker -- DeepSORT \cite{Wojke2017}. Once we have obtained a trajectory of boxes for each person, we can now extract the motion feature. Specifically, we first extract the optical flow for each box trajectory, and then further decompose it into smaller temporal super-pixels using \cite{Chang2013}. The reason for decomposition is that we believe the inertial data measured by the smartphone is only correlated with a part of the body where the smartphone is held, instead of the entire body. Without this decomposition, the optical flow representing the motion of the entire body might not be easily matched with the inertial feature representing the motion of a part of the body, thus leading to inferior localization performance.

\begin{figure}[t]
  \centering
  \includegraphics[width=1.0\columnwidth]{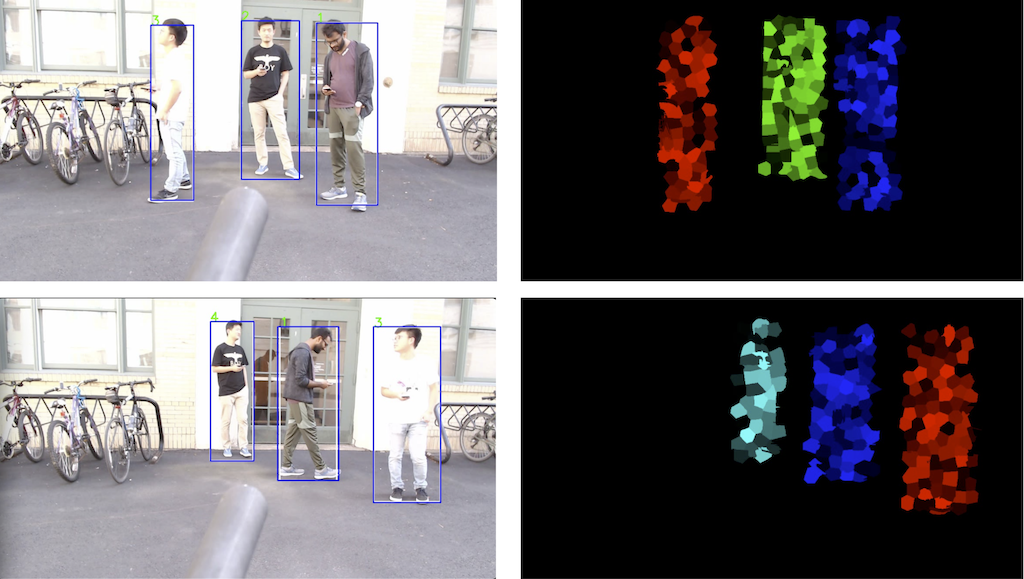}
  \vspace{-0.7cm}
  \caption{\textbf{(Left)} YOLOv3 person detections. \textbf{(Right)} Temporal super-pixels (TSP) for each tracked person in the video. Average optical flow is computed as the motion feature for each TSP representing different body parts.}
  \label{fig:tsp}
  \vspace{-0.4cm}
\end{figure}
   
Formally speaking, given a video segment $V_{t:t+T}$ with $T$ frames, we denote the set of temporal super-pixels (TSPs) as $\boldsymbol{\xi} = \{ \xi_1,\xi_2,\ldots,\}$. We then filter out the TSPs that do not lie within the trajectories of boxes and obtain a subset of TSPs denoted as $\boldsymbol{\xi}^n \subset \boldsymbol{\xi}$ for each person. To obtain the motion feature for each TSP, we compute the average optical flow over all pixels for each temporal slice of a TSP: $$\mathbf{v}_{\xi^n_i} = [(dx_t, dy_t), (dx_{t+1}, dy_{t+1}), ..., (dx_{t+T-1}, dy_{t+T-1})],$$ where each vector $\mathbf{v}_{\xi^n_i}$ represents the motion of a part of the human body as shown in Fig. \ref{fig:tsp}. 

Although the TSP features are sufficient to represent the motion information of different body parts in the video, there is still a gap between the TSP features and the inertial 3D motion features as the TSP features are computed in the 2D image space, \emph{i.e.}, the perspective projection of the person's 3D motion. To alleviate this issue and bridge the gap between the 2D and 3D space, we include extra information that is related to the 3D depth and orientation of the person, which can implicitly help the matching between the learned visual and inertial feature embeddings. Specifically, we use two types of information obtained from the video segment:
\begin{enumerate}
    \item The height and width of the person as an indication of the distance between the person and the camera: $$\mathbf{b}^n = [(h_t, w_t), (h_{t+1}, w_{t+1}), ..., (h_{t+T-1}, w_{t+T-1})].$$
    \item The relative positions of the person's left and right shoulder keypoints to the bounding box center as an indication of the body orientation relative to the camera: $$\mathbf{k}^n = [(\mathbf{ls}_t, \mathbf{rs}_t), (\mathbf{ls}_{t+1}, \mathbf{rs}_{t+1}), (\mathbf{ls}_{t+T-1}, \mathbf{rs}_{t+T-1})],$$ where $\mathbf{ls}$ and $\mathbf{rs}$ are tuples of the keypoint's $x$ and $y$ coordinates relative to the box center's coordinate in the image frame. We choose the shoulder keypoints because their positions are stable to the body orientation.
\end{enumerate} 

We use the bounding box trajectory outputs obtained by YOLOv3 and DeepSORT to extract the width and height information for each person. To obtain the positions of shoulder keypoints, we first use AlphaPose \cite{xiu2018poseflow} to detect $17$ keypoints of the full body, and then only select the two points representing the shoulder joints. If a person is temporarily occluded, we apply linear interpolation on the bounding box and keypoint trajectories. If a person exits out of the camera frame, we pad zeros to the trajectories.

\subsection{Inertial Feature Extraction}

To match with the visual motion feature, we also need to extract an inertial feature, which represents the 3D motion of the smartphone for the target person. Given the IMU data containing the 3D linear acceleration (pre-processed by removing the affect of gravity) $\mathbf{a} = [\vec{a}_x, \vec{a}_y, \vec{a}_z]$ and angular velocity $\boldsymbol{\omega} = [\vec{\omega}_x, \vec{\omega}_y, \vec{\omega}_z]$ in the smartphone's local coordinate frame, we construct the inertial feature for target person $n$ denoted as $g_{\textrm{IMU}}^n = [\vec{a}_x, \vec{a}_y, \vec{a}_z , \vec{\omega}_x, \vec{\omega}_y, \vec{\omega}_z]^T$ by concatenating linear acceleration and angular velocity. As a result, the inertial feature $g_{\textrm{IMU}}^n$ is a $6\times T'$ matrix where $T'$ is the number of frames temporally aligned with the video segment's time window. As the IMU frame rate is $100$Hz, with a ratio of 3.33:1 to the video frame rate of $30$Hz, we uniformly sample the inertial frames so that $T' = 3\times T$, where $T$ is the number of frames in a video segment. 

\begin{figure*}[t]
\centering
\includegraphics[width=\textwidth]{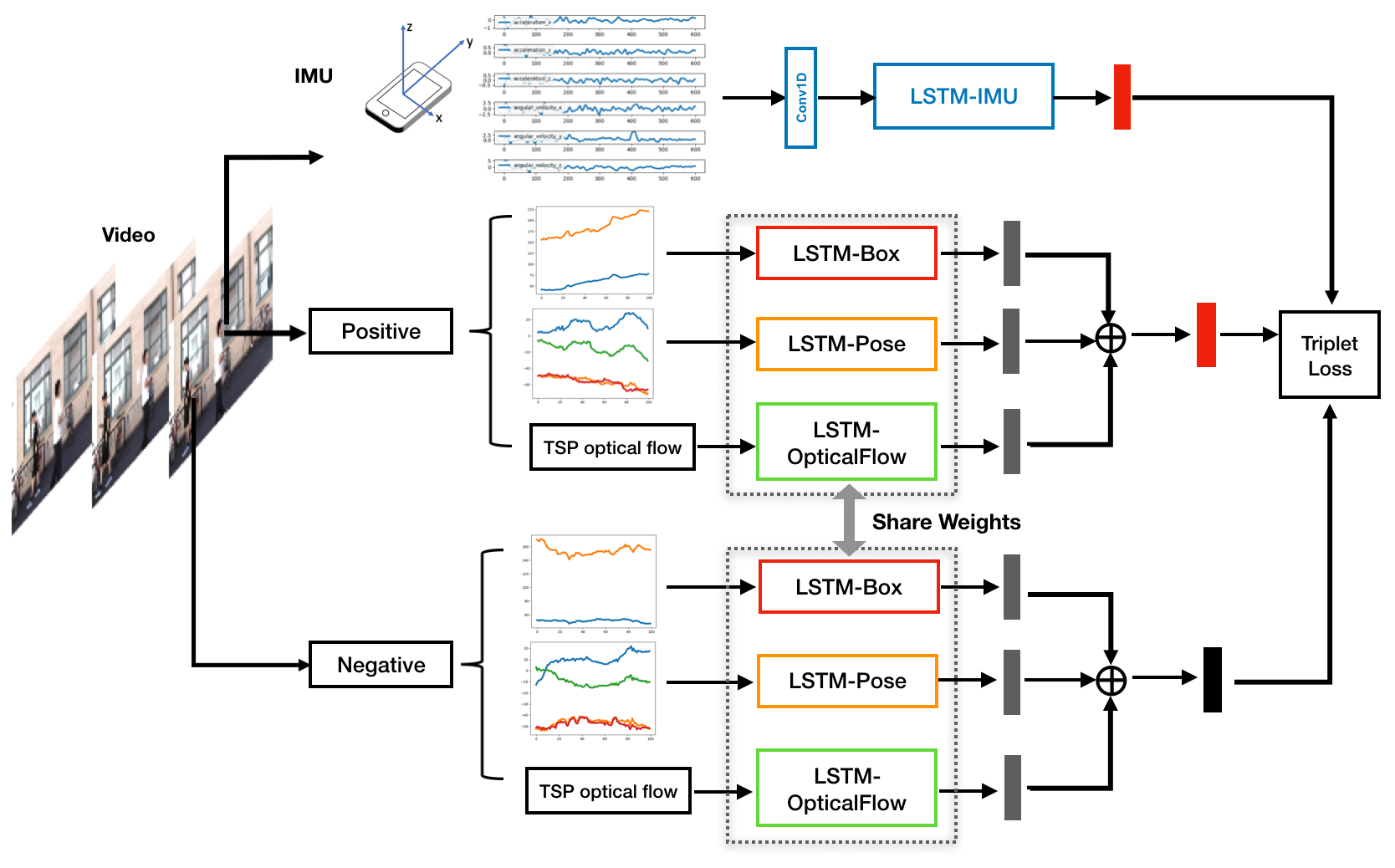}
\vspace{-1cm}
\caption{\textbf{Proposed Network}. Our network has one branch to extract the inertial feature of the target person and two branches to extract the visual features from one positive and one negative sample. At each iteration of training, the positive visual feature is extracted from the target person while the negative visual feature is from a randomly picked different person. Once the raw inertial and visual features are extracted, they are fed into our visual-inertial feature transformer so that the transformed feature embeddings lie in a same feature space. A triplet loss is then applied to minimize the L2 distance between the inertial embedding and the positive visual embedding while maximize the L2 distance between the inertial embedding and the negative visual embedding. At test time, we compute the visual embeddings for all persons in the video and also compute the inertial embedding of the target person. The predicted target person in the video is then the person whose visual embedding has minimum distance to the target person's inertial embedding.}
\label{fig:model}
\vspace{-0.4cm}
\end{figure*}

\subsection{Learning the Joint Visual-Inertial Feature Space}

Although the raw visual and inertial feature contain sufficient information representing the person's 3D motion, they still lie in different feature spaces as they are obtained from different source of data and thus it is difficult to directly match them. To overcome this issue, we propose to learn a feature transformer that further transform the raw visual and inertial features into a joint feature space so that the matching for a same person is possible.

The proposed network for learning the joint feature space is shown in Fig. \ref{fig:model}. To transform the raw visual feature into the joint space while model the temporal dependency, we first apply three LSTM networks for the TSP features (\textcolor{Green}{green}), bounding box size data (\textcolor{red}{red}) and pose keypoints data (\textcolor{orange}{orange}) respectively, each with different weights. Then, we combine these three features together as the visual feature. To transform the raw inertial feature into the joint space, we first use a 1D convolution layer to reduce the dimensionalities of the inertial feature to be the same as the visual feature. Then, we also apply an LSTM network (\textcolor{blue}{blue}) to model the temporal dependency for the inertial feature. For both visual and inertial features, we use the hidden state from the LSTM at each timestep as the final output embedding, which are formally defined as follows:
\begin{align*}
\resizebox{\hsize}{!}{
$H_{\textrm{VIS}}(\mathbf{v}_{\xi^n_i}, \mathbf{b}^n, \mathbf{k}^n) = f_{\text{OF}}(\mathbf{v}_{\xi^n_i}) + \alpha f_{\text{Pose}}(\mathbf{k}^n) + \beta f_{\text{Box}}(\mathbf{b}^n)$,
}
\end{align*}
\vspace{-0.7cm}
\begin{align*}
H_{\textrm{IMU}}(g_{\textrm{IMU}}^n) = f_{\text{IMU}}(g_{\textrm{IMU}}^n),
\end{align*}
where the final visual embedding is computed by summing over three different input features and $\alpha$ and $\beta$ are two hyper-parameters defining the weights. As each person $n$ has a set of TSPs $\boldsymbol{\xi}^n$ and thus we have $|\boldsymbol{\xi}^n|$ final visual embeddings, we duplicate the number of final inertial embeddings so that we have the same number of visual and inertial embeddings for each person in a time window with $T$ frames. During training, we use every pair of the inertial and visual embeddings and minimizing the L2 distance between them if they belong to the same person:
\begin{align*}
\resizebox{\hsize}{!}{
$\mathcal{L}(g_{\textrm{IMU}}^n, g_{\textrm{VIS}}^n(\xi_i)) = ||H_{\text{VIS}}(g_{\textrm{VIS}}^n(\xi_i)) - H_{\text{IMU}}(g_{\textrm{IMU}}^n)||_2$,
}
\end{align*}
where $g^n_{\textrm{VIS}}(\xi_i)$ denotes the tuple $(\mathbf{v}_{\xi^n_i}, \mathbf{b}^n, \mathbf{k}^n)$. Additionally, we use the triplet loss as in \cite{hoffer2015, hermans2017}. Specifically, for each target person $n$ with the inertial embedding, we use the visual embedding obtained from the same target person as a positive example and use the visual embedding obtained from a randomly sampled different person as a negative example. The positive and negative samples share the same weights in the LSTM networks (\emph{i.e.}, LSTM-OpticalFlow, LSTM-Pose, LSTM-Box). Then, the triplet loss is applied to minimize the L2 distance between the inertial and positive visual embedding and maximize the L2 distance between the inertial and negative visual embedding:
\begin{align*}
\resizebox{\hsize}{!}{
$\begin{aligned}
\mathcal{L}(g_{\textrm{IMU}}^n, g_{\textrm{VIS}}^+(\xi_i), g_{\textrm{VIS}}^-(\xi_j)) = \max(&||H_{\text{VIS}}(g_{\textrm{VIS}}^+(\xi_i)) - H_{\text{IMU}}(g_{\textrm{IMU}}^n)||_2 -\\ &||H_{\text{VIS}}(g_{\textrm{VIS}}^-(\xi_j)) - H_{\text{IMU}}(g_{\textrm{IMU}}^n)||_2 + \kappa, 0),
\end{aligned}$
}
\end{align*}
where $g_{\textrm{VIS}}^+(\xi_i)$ is the extracted visual feature given a TSP from the positive person and $g_{\textrm{VIS}}^-(\xi_j)$ is the extracted visual feature given a randomly selected TSP from a non-target person. $\kappa$ is the margin separating the positive and negative feature space. At test time, given a video segment $V_{t:t+T}$ with $N$ people in the scene, we choose one person as the target person at a time and compute its inertial embedding. Meanwhile, we compute the visual embedding for all persons in the video. Then, the predicted target person is the person whose visual embedding averaged over all TSPs has the minimum distance to the target person's inertial embedding:
\begin{align*}
\resizebox{\hsize}{!}{
$\hat{n} = \arg\min_{n' \in [N]} \frac{1}{|\boldsymbol{\xi}^{n'}|}\sum_{i=1}^{|\boldsymbol{\xi}^{n'}|}||H_{\text{VIS}}(g_{\textrm{VIS}}^{n'}(\xi_i)) - H_{\text{IMU}}(g_{\textrm{IMU}}^n)||_2$,
}
\end{align*}
where $|\boldsymbol{\xi}^{n'}|$ is the number of TSP's for person $n'$.

\section{A New Visual-Inertial Dataset}

To train our proposed method for visual-inertial person localization in the wild, we need a dataset with synchronized video and inertial data that include multiple people acting freely outside, each carrying a smartphone in their hand. However, existing visual-inertial datasets \cite{Henschel2019, DeLaTorre2008, Trumble2017} do not satisfy these requirements and often have three limitations: 1) they rigidly attach the inertial sensor to person's body (\emph{e.g.}, limb or back) so that the motion of the inertial sensor tightly aligns with the body part; 2) they often record the data in the indoor setting; 3) only one person is recorded at one time. As a result, prior datasets are not applicable to our challenging visual-inertial person localization task, and we have to collect a new dataset to satisfy the task conditions, which we plan to make public to encourage future research on similar tasks. 

\begin{table}
    \centering
    \caption{Statistics of the video data collected in our dataset.}
    \vspace{-0.3cm}
    \resizebox{\columnwidth}{!}{
    \begin{tabular}{@{}lrrrrr@{}}
        \toprule
        Number of people                & 2      & 3      & 4      & 5      & 6     \\
        Number of videos                & 17     & 15     & 11     & 7      & 8     \\
        Number of total frames          & 12,900 & 19,600 & 21,400 & 10,084 & 5,000 \\
        \bottomrule
    \end{tabular}
    }
    \label{tab:data}
    \vspace{-0.5cm}
\end{table}

\subsection{Video Recording}
We set up a static HD webcam with a resolution of $1920\times1080$ on a tripod about one meter above the ground for video recording, similar to the setting of a dashboard camera in a car. We choose to record the video outside of public buildings in the daytime, in order to mimic real-world autonomous vehicle pickup scenarios. At each time of the recording, we hire 2 to 6 different volunteers, assign a smartphone to each of the volunteer, and ask the volunteers to perform casual random motion (\emph{e.g.}, walking or standing still while holding the smartphones at hands). Each video recording is about half to two minutes long with a frame rate of $30$Hz. In total, we have recorded $58$ videos with a total of $68984$ frames. We summarize the statistics of our data recording in Table \ref{tab:data}. Our dataset contains common types of pedestrian motion such as standing, walking and turning, recorded at four different backgrounds to increase the diversity of the dataset. Also, we do not provide and allow to use the calibration parameters of the camera in our dataset, as in the real world the calibration parameters of the dashboard camera might vary across vehicles and not available to our approach for person localization. Each video frame is time-stamped with the UTC time for synchronization with IMU. 

\subsection{IMU Recording}
We use Apple iPhone (model 7 and 8) as the smartphone device to collect the inertial data. To that end, we have developed an iOS application with the iOS Core Motion Framework to obtain the linear acceleration and angular velocity data from the onboard accelerometer and gyroscope. For linear acceleration, we use the processed data by the device that only reflects the user-generated acceleration after removing the gravity. The IMU data is recorded at $100$Hz with UTC timestamps. At each time of the recording, we ask the volunteers to start the iOS application on their smartphones so that the data can be saved to the device. As the data synchronization is handled by matching the timestamp, volunteers do not need to start the application exactly at the same time.

\subsection{Data Pre-Processing}
As optical flow is needed to obtain the visual embedding, we pre-compute the flow for all videos so that the online training can be faster. However, computing optical flow on the raw images with a resolution of $1920 \times 1080$ is very expensive, we thus downsample the raw images to a resolution of $691\times389$ to speed up the pre-processing step. Also, as our network can only process a short video segment at a time, we convert the raw video and inertial data into short segments using a sliding window approach. Specifically, we use a window size of $150$ frames (the best experimental setting) and a step size of $20$ frames, which results in over $3,000$ synchronized video and inertial data segments. 

As we have the inertial data for all persons in each data segment, we can iteratively mark each person in the data segment as the target person. This means that each video segment can serve as $m$ data segment samples during training and evaluation where $m$ equals to the number of persons in the video. This data augmentation technique further increases the number of our data segment samples about four times. Additionally, As our proposed method relies on the motion feature matching between the inertial and visual modalities, it is difficult to perform the matching if the target person has nearly no motion. As a result, we filter out data segment samples (about $25\%$ of the data) during training and evaluation where the target person's IMU acceleration magnitude has a standard deviation less than $0.02 m/s^2$. In the future, we will deal with this limitation with additional features that are more sensitive to small motions and achieve person localization even the target person has no motion. 

\section{Experiments}

\begin{figure*}[t]
\centering
\includegraphics[width=\textwidth]{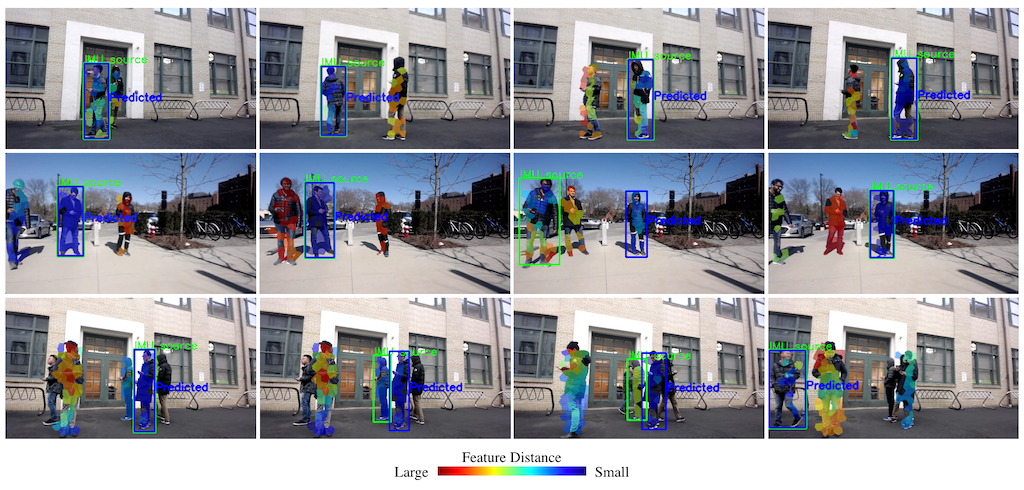}
\vspace{-0.9cm}
\caption{We show qualitative results of our method for visual-inertial person localization on three test videos with different number of people in the scene. The green box indicated as the IMU source is the target person while the blue box is the predicted target person by our method. When the green and blue boxes fall on a same person, it is a correct match. We show both successful and failure cases in the results. Also, we visualize the distance of the visual feature for each TSP to the inertial feature of the true target person.}
\label{fig:visual}
\vspace{-0.3cm}
\end{figure*}

\subsection{Evaluation Details}
Since our visual-inertial person localization is formalized as a matching problem, we use the classification rate as our evaluation metric, namely the probability that our method can output a correct match for the target person. We split our collected data into train (35 videos), validation (2 videos) and test set (4 videos), where each set contains videos with different number of people. Note that some challenging videos are not used, \emph{e.g.}, videos with moving cameras, as our current method is limited to deal with static video. The evaluation of our method and baselines is only conducted on the test set, while the validation set is used for parameter tuning. Usually, when there are more people in the scene, it is more likely that people will have similar motion (\emph{e.g.}, walking in the same direction), which makes the data more difficult for matching and localization. Naturally, we expect that our visual-inertial person localization task to become harder when there are more people in the scene. We show quantitative results with $\{2, 3, 4, 5\}$ people in the scene.

\subsection{Comparison to Baseline Methods}
As there is not open-source code released for \cite{Henschel2019}, we try our best to re-implement the module proposed in \cite{Henschel2019} for visual-inertial person localization. Besides \cite{Henschel2019}, there is no other baseline in prior work to compare against, we thus devise several competitive baselines based on the existing techniques. For fair comparison, we use the same time window of $K=150$ frames for our method. All the baselines are listed below (1-4 are direct feature matching using cosine distance; 5-6 are supervised learning for transforming one modality with the other being the ground-truth label. At test time, the predicted person has the minimum distance between her visual feature and the query inertial feature):

\begin{enumerate}[leftmargin=*]\itemsep0.1cm 
    \item \textbf{Velocity Magnitude.} For the visual feature, we compute a sequence of magnitude of the optical flow for each TSP feature: $\{\sqrt{v_x^2 + v_y^2} \}_{k=1}^{K}$. For the IMU feature, we compute the 3D velocity integrated from the 3D linear acceleration: $\vec{v}_t=\vec{v}_{t-1}+ \vec{a}_t \nabla t$. Then, we also compute a sequence of velocity magnitude: $\{||\vec{v}_t||_2\}_{k=1}^{K}$, which is used to match with the visual velocity magnitude.
    
    \item \textbf{Acceleration Magnitude.} We compute a sequence of magnitude of the optical flow gradients for each TSP feature: $\{ \sqrt{a_x^2 + a_y^2} \}_{k=1}^{K}$, representing visual acceleration magnitude. For the IMU feature, we compute a sequence of magnitude of the linear acceleration: $\{||\vec{a}_t||_2\}_{k=1}^{K}$. Then, we match two computed acceleration magnitudes for visual-inertial person localization.
    
    \item \textbf{Velocity Magnitude Histogram.} We first use the same method as 1) to compute the velocity magnitude. Then, the sequence of velocity magnitude is binned to create a velocity magnitude histogram, where we use $150$ bins.
    
    \item \textbf{Acceleration Magnitude Histogram.} We first use the same method as 2) to compute the acceleration magnitude. Then, the sequence of acceleration magnitude is binned to create a histogram, where we use $150$ bins.
    
    \item \textbf{3D Orientation.} We re-implement the image-based 3D orientation estimation technique in \cite{Henschel2019} where the person's 3D orientation is predicted from a VGG16 network with RGB image input. Following \cite{Henschel2019}, we add two fully connected layers to the VGG16 backbone and learn the mapping from image to the person's 3D orientation. The network employs two adjacent images of a person in the tracklet as the input and regresses the 3D orientation change. We train the network using the angular velocity obtained from the IMU as ground truth. At test time, we can obtain a sequence of 3D orientation change for the person from the image, in order to match with the 3D angular velocity obtained from the inertial data.
    
    \item \textbf{2D Optical Flow.} For the visual feature, we use the optical flow for each TSP feature. Meanwhile, we learn to map a sequence of 3D acceleration $\vec{a}$ and 3D angular velocity $\vec{w}$ to a sequence of velocity in the 2D space. The mapping function is learned by supervised learning where the input is $(\vec{a},\vec{\omega})$ and the ground truth is the 2D optical flow. We use the same conv-1D and LSTM-IMU network in our method as the mapping function. At test time, we match the 2D optical flow from the visual feature with the estimated 2D velocity from the inertial feature.
\end{enumerate}

\begin{table}[t]
\centering
\caption{Quantitative comparison of accuracy on test videos with different number of people.}
\vspace{-0.3cm}
\resizebox{\columnwidth}{!}{
\begin{tabular}{@{}lrrrrr@{}}
    \toprule
    Method                                 & N=2   & N=3   & N=4   & N=5   \\
    \midrule
    Random Guess                           & 0.500 & 0.333 & 0.250 & 0.200 \\
    \midrule
    1) Velocity Magnitude                  & 0.500 & 0.379 & 0.429 & 0.456 \\
    2) Velocity Mag. Histogram             & 0.500 & 0.379 & 0.464 & 0.474 \\
    3) Acceleration Magnitude              & 0.500 & 0.379 & 0.536 & 0.456 \\
    4) Accel. Mag. Histogram               & 0.500 & 0.379 & 0.429 & 0.456 \\
    5) 3D Orientation \cite{Henschel2019}  & 0.502 & 0.344 & 0.306 & 0.194 \\
    6) 2D Optical Flow                     & 0.682 & 0.402 & 0.392 & 0.439 \\
    \midrule
    \textbf{Ours} & \textbf{0.906} & \textbf{0.840} & \textbf{0.667} & \textbf{0.816}\\
    \bottomrule
\end{tabular}}
\label{tab:baselines}
\vspace{-0.3cm}
\end{table}

\vspace{0.1cm}
We show quantitative comparison of our method and above baselines in Table \ref{tab:baselines}. We can see that baseline methods 1 to 4 with hand-designed features often perform poorly as the motion features from the visual and inertial modalities are in different feature spaces, and it is challenging to directly match them. Also, learning to transform one modality to the other (\emph{i.e.}, baseline methods 5 and 6) does not achieve superior performance. This proves again the significant gap between the two modalities. We show that, only when we transform the features from both two modalities into a joint feature space in our method, significant improvement can be achieved across videos with different number of people in the scene. Moreover, we noticed that our method performs better when $N$=$5$ than $N$=$4$, which is counter-intuitive. To find out the reasons, we visualized the results and found that the test videos with 5 people happen to have more distinct motions among the candidates compared to the test videos with 4 people, which proves that motion diversity in the data is a key factor to our method's performance. 

Additionally, we show qualitative results of our method on the test set in Fig. \ref{fig:visual}. The results show that our method can predict a correct match in most of the frames, while in the failure cases the true target is often confused with the false predicted target with similar motion (best viewed in video).

\begin{table}
\centering
\caption{Performance of our method on test videos with respect to different window lengths.}
\vspace{-0.3cm}
\resizebox{\columnwidth}{!}{
\begin{tabular}{@{}lrrrrr@{}}
   \toprule
   Window Length / frames & N=2   & N=3   & N=4   & N=5   \\
   \midrule
   100 & 0.820 & 0.747 & 0.643 & 0.456\\
   150 (\textbf{Ours}) & \textbf{0.906} & \textbf{0.840} & \textbf{0.667} & \textbf{0.816}\\
   180 & 0.667 & 0.447 & 0.333 & 0.429\\
   200 & 0.556 & 0.631 & 0.605 & 0.480\\
   \bottomrule
\end{tabular}}
\label{tab:window}
\vspace{-0.6cm}
\end{table}

\subsection{Ablation Study: Length of the Time Window}
As more discriminative motion feature can be found in a longer time window, we believe the length of time window is an important factor to the performance of our method and run ablation experiments with respect to it. Specifically, we run experiments with a window length of $100, 150, 180, 200$ frames (\emph{i.e.}, $3.33, 5, 6, 6.67$ seconds). We use the same step size of $20$ frames ($0.67$ seconds) as the sliding time window for all experiments. We found that the highest accuracy is achieved with a window length of $150$ frames. Also, we observed a performance drop when the window length goes beyond $150$ frames. It turns out that when the window length increases beyond $150$ frames, the number of data samples drops significantly as most of the person trajectories in our dataset are short due to heavy occlusion by other persons. As a result, due to limited data samples, training process of our network becomes unstable and evaluation is not trustable. Additionally, a longer time window means a larger latency of our method. Therefore, we did not further investigate longer time window but use the window of $150$ frames in our model.

\begin{table}
\centering
\caption{Performance of our method with respect to different variations of the inertial feature representation.}
\vspace{-0.3cm}
\resizebox{\columnwidth}{!}{
\begin{tabular}{@{}lrrrr@{}}
    \toprule
    Inertial Feature Representation                         & N=2   & N=3   & N=4   & N=5   \\
    \midrule   
    $(\hat{\mathbf{v}}, \mathbf{a}, \boldsymbol{\omega})$   & 0.680 & 0.793 & 0.357 & 0.509 \\
    $(\mathbf{a}, \boldsymbol{\omega})$                     & 0.820 & 0.747 & 0.643 & 0.403 \\
    $(\hat{\mathbf{v}}, \boldsymbol{\omega})$               & 0.600 & 0.632 & 0.321 & 0.491 \\
    $(\mathbf{a}_{\text{LPF}}, \boldsymbol{\omega}_{\text{LPF}})$ (\textbf{Ours}) & \textbf{0.906} & \textbf{0.840} & \textbf{0.667} & \textbf{0.816}\\
    \bottomrule
\end{tabular}}
\label{tab:imu}
\vspace{-0.3cm}
\end{table}

\subsection{Ablation Study: Inertial Feature Representation}
The use of a different feature representation can result in significant differences in performance. Here, we first investigate different variations of the inertial feature representation. In addition to the linear acceleration and angular velocity, we believe the linear velocity might be also an informative feature for matching with the visual motion feature. To that end, we integrate the linear acceleration from the IMU to estimate the linear velocity $\hat{\mathbf{v}} = [\hat{v}_x, \hat{v}_y, \hat{v}_z]$ as an additional 3D motion information. As we ask the volunteers to stand still at the beginning of each video recording and then to start moving freely, we can use an initial velocity of $0$ for the integration. Results in Table \ref{tab:imu} first row ($\hat{\mathbf{v}}$, $\mathbf{a}$, $\boldsymbol{\omega}$) show that concatenating the estimated linear velocity with the linear acceleration and angular velocity unfortunately performs slightly worse than without adding the linear velocity as shown in the second row of Table \ref{tab:imu}. Also, we experiment a variant that concatenates the estimated linear velocity and angular velocity in the third row of Table \ref{tab:imu}, which has a even lower performance than both the first and second row. These results demonstrate that the estimated linear velocity through integration might not be accurate enough due to the error accumulation from the IMU drift and thus we do not use the linear velocity in our final model.

Additionally, as the inertial data obtained from the IMU sensor often has high-frequency noise, we experiment the effect of a low-pass filter to our method. Specifically, we apply the filter to both the linear acceleration and angular velocity and obtain a smoother version of the inertial features $(\mathbf{a}_{\text{LPF}}, \boldsymbol{\omega}_{\text{LPF}})$, which turns out improving overall performance by $11.5\%$ across settings with different number of people.

\begin{table}
\centering
\caption{Performance of our method with respect to the loss weights on the keypoint and bounding box size features.}
\vspace{-0.3cm}
\resizebox{\columnwidth}{!}{
\begin{tabular}{@{}lrrrrr@{}}
    \toprule
    Loss Weight $\alpha$ & 0.0 & 0.2 & 0.5 & 0.8 & 1.0  \\
    N=2     & 0.875 & 0.813 & \textbf{0.906} & \textbf{0.906} & 0.750\\
    N=3     & 0.671 & 0.780 & \textbf{0.840} & 0.758 & 0.597\\
    \midrule
    Loss Weight $\beta$ & 0.0 & 0.2 & 0.5 & 0.8 & 1.0  \\
    N=2     & 0.700 & \textbf{0.906} & 0.760 & 0.860 & 0.667\\
    N=3     & 0.563 & \textbf{0.840} & 0.598 & 0.701 & 0.632\\
    \bottomrule
\end{tabular}}
\label{tab:features}
\vspace{-0.6cm}
\end{table}

\subsection{Ablation Study: Visual Feature Representation}
To verify whether adding the relative positions of person's shoulder keypoints and the bounding box size to the visual feature is useful in our model, we run experiments with different values of the hyper-parameters $\alpha$ and $\beta$, which controls how much we use the information of the shoulder keypoints and bounding box size during training. For example, when $\alpha$ or $\beta$ are $0$, we turn off the branch for learning shoulder keypoint and bounding box size features. From the results in Table \ref{tab:features}, we observed that the shoulder keypoint and bounding box size features are indeed useful with proper weights and improve the performance of our method on test videos with different number of people. 

\section{CONCLUSIONS}

We explored the possibility of using the inertial data to localize the target person in the video, in the case where we do not have access to the target person's appearance information in advance. We term this proposed task as the visual-inertial person localization. To solve this task, we first collected a new large visual-inertial dataset, which is significantly different from existing datasets in that our new dataset contains multiple people in the wild and does not have strict constraint on the attached location of the inertial sensor. Additionally, we proposed an effective approach that learns a transformer and maps the visual and inertial features into a joint feature space for matching. Through extensive experiments, we showed effectiveness of each component of our method and demonstrated that the proposed method outperforms competitive baselines on our challenging dataset.

\section*{ACKNOWLEDGMENT}

This work is sponsored in part by Highmark.

\bibliographystyle{IEEEtran}
\bibliography{IEEEabrv,main}

\begin{thebibliography}{10}
\providecommand{\url}[1]{#1}
\csname url@rmstyle\endcsname
\providecommand{\newblock}{\relax}
\providecommand{\bibinfo}[2]{#2}
\providecommand\BIBentrySTDinterwordspacing{\spaceskip=0pt\relax}
\providecommand\BIBentryALTinterwordstretchfactor{4}
\providecommand\BIBentryALTinterwordspacing{\spaceskip=\fontdimen2\font plus
\BIBentryALTinterwordstretchfactor\fontdimen3\font minus
  \fontdimen4\font\relax}
\providecommand\BIBforeignlanguage[2]{{%
\expandafter\ifx\csname l@#1\endcsname\relax
\typeout{** WARNING: IEEEtran.bst: No hyphenation pattern has been}%
\typeout{** loaded for the language `#1'. Using the pattern for}%
\typeout{** the default language instead.}%
\else
\language=\csname l@#1\endcsname
\fi
#2}}

\bibitem{Kayukawa2019}
S.~Kayukawa, K.~Higuchi, J.~Guerreiro, S.~Morishima, Y.~Sato, K.~Kitani, and
  C.~Asakawa, ``{BBeep: A Sonic Collision Avoidance System for Blind Travellers
  and Nearby Pedestrians},'' \emph{CHI}, 2019.

\bibitem{Manglik2019}
A.~Manglik, X.~Weng, E.~Ohn-bar, and K.~M. Kitani, ``{Forecasting
  Time-to-Collision from Monocular Video: Feasibility, Dataset, and
  Challenges},'' \emph{IROS}, 2019.

\bibitem{Guerreiro2019}
J.~Guerreiro, D.~Sato, S.~Asakawa, H.~Dong, K.~M. Kitani, and C.~Asakawa,
  ``{CaBot: Designing and Evaluating an Autonomous Navigation Robot for Blind
  People},'' \emph{ASSETS}, 2019.

\bibitem{Wojke2017}
N.~Wojke, A.~Bewley, and D.~Paulus, ``{Simple Online and Realtime Tracking with
  a Deep Association Metric},'' \emph{ICIP}, 2017.

\bibitem{Bewley2016}
A.~Bewley, Z.~Ge, L.~Ott, F.~Ramos, and B.~Upcroft, ``{Simple Online and
  Realtime Tracking},'' \emph{ICIP}, 2016.

\bibitem{Weng2019_3dmot}
X.~Weng, J.~Wang, D.~Held, and K.~Kitani, ``{3D Multi-Object Tracking: A
  Baseline and New Evaluation Metrics},'' \emph{IROS}, 2020.

\bibitem{Kiran2020}
B.~R. Kiran, I.~Sobh, V.~Talpaert, P.~Mannion, A.~A.~A. Sallab, S.~Yogamani,
  and P.~P{\'{e}}rez, ``{Deep Reinforcement Learning for Autonomous Driving: A
  Survey},'' \emph{arXiv:2002.00444}, 2020.

\bibitem{Wang2018}
S.~Wang, D.~Jia, and X.~Weng, ``{Deep Reinforcement Learning for Autonomous
  Driving},'' \emph{arXiv:1811.11329}, 2018.

\bibitem{Weng2020}
X.~Weng, Y.~Yuan, and K.~Kitani, ``{Joint 3D Tracking and Forecasting with
  Graph Neural Network and Diversity Sampling},'' \emph{arXiv:2003.07847},
  2020.

\bibitem{Leon2019}
F.~Leon and M.~Gavrilescu, ``{A Review of Tracking, Prediction and Decision
  Making Methods for Autonomous Driving},'' \emph{arXiv:1909.07707}, 2019.

\bibitem{Man2019}
Y.~Man, X.~Weng, X.~Li, and K.~Kitani, ``{GroundNet: Monocular Ground Plane
  Normal Estimation with Geometric Consistency},'' \emph{ACMMM}, 2019.

\bibitem{Luo2018}
W.~Luo, B.~Yang, and R.~Urtasun, ``{Fast and Furious: Real Time End-to-End 3D
  Detection, Tracking and Motion Forecasting with a Single Convolutional
  Net},'' \emph{CVPR}, 2018.

\bibitem{Zeng2019}
W.~Zeng, W.~Luo, S.~Suo, A.~Sadat, B.~Yang, S.~Casas, and R.~Urtasun,
  ``{End-to-End Interpretable Neural Motion Planner},'' \emph{CVPR}, 2019.

\bibitem{Weng20202}
X.~Weng, J.~Wang, S.~Levine, K.~Kitani, and N.~Rhinehart, ``{Sequential
  Forecasting of 100,000 Points},'' \emph{arXiv:2003.08376}, 2020.

\bibitem{Rangesh2019}
A.~Rangesh and M.~M. Trivedi, ``{No Blind Spots: Full-Surround Multi-Object
  Tracking for Autonomous Vehicles using Cameras {\&} LiDARs},'' \emph{IV},
  2019.

\bibitem{Yurtsever2019}
E.~Yurtsever, J.~Lambert, A.~Carballo, and K.~Takeda, ``{A Survey of Autonomous
  Driving: Common Practices and Emerging Technologies},''
  \emph{arXiv:1906.05113}, 2019.

\bibitem{Weng20203}
X.~Weng, Y.~Wang, Y.~Man, and K.~Kitani, ``{GNN3DMOT: Graph Neural Network for
  3D Multi-Object Tracking with 2D-3D Multi-Feature Learning},'' \emph{CVPR},
  2020.

\bibitem{Badue2019}
C.~Badue, R.~Guidolini, R.~V. Carneiro, P.~Azevedo, V.~B. Cardoso, A.~Forechi,
  L.~Jesus, R.~Berriel, T.~Paix{\~{a}}o, F.~Mutz, L.~Veronese,
  T.~Oliveira-Santos, and A.~F. {De Souza}, ``{Self-Driving Cars: A Survey},''
  \emph{arXiv:1901.04407}, 2019.

\bibitem{Weng2019_mono3d}
X.~Weng and K.~Kitani, ``{Monocular 3D Object Detection with Pseudo-LiDAR Point
  Cloud},'' \emph{ICCVW}, 2019.

\bibitem{Wang2020_gnndettrk}
Y.~Wang, X.~Weng, and K.~Kitani, ``{Joint Detection and Multi-Object Tracking
  with Graph Neural Networks},'' \emph{arXiv:2006.13164}, 2020.

\bibitem{Chang2013}
J.~Chang, D.~Wei, and J.~W. Fisher, III, ``{A Video Representation Using
  Temporal Superpixels},'' \emph{CVPR}, 2013.

\bibitem{Hochreiter1997}
S.~Hochreiter and J.~{Urgen Schmidhuber}, ``{Long Short-Term Memory},''
  \emph{Neural Computation}, 1997.

\bibitem{schroff2015}
F.~Schroff, D.~Kalenichenko, and J.~Philbin, ``{FaceNet: A Unified Embedding
  for Face Recognition and Clustering},'' \emph{CVPR}, 2015.

\bibitem{Henschel2019}
R.~Henschel, T.~von Marcard, and B.~Rosenhahn, ``{Simultaneous Identification
  and Tracking of Multiple People Using Video and IMUs},'' \emph{CVPRW}, 2019.

\bibitem{Marcard2018}
T.~von Marcard, R.~Henschel, M.~J. Black, B.~Rosenhahn, and G.~Pons-Moll,
  ``{Recovering Accurate 3D Human Pose in the Wild Using IMUs and a Moving
  Camera},'' \emph{ECCV}, 2018.

\bibitem{DIP2018}
Y.~Huang, M.~Kaufmann, E.~Aksan, M.~J. Black, O.~Hilliges, and G.~Pons-Moll,
  ``{Deep Inertial Poser: Learning to Reconstruct Human Pose from Sparse
  Inertial Measurements in Real Time},'' \emph{SIGGRAPH Asia}, 2018.

\bibitem{SIP2017}
T.~{von Marcard}, B.~Rosenhahn, M.~Black, and G.~Pons-Moll, ``{Sparse Inertial
  Poser: Automatic 3D Human Pose Estimation from Sparse IMUs},''
  \emph{Eurographics}, 2017.

\bibitem{DeLaTorre2008}
F.~de~la Torre, J.~K. Hodgins, J.~Montano, and S.~Valcarcel, ``{Detailed Human
  Data Acquisition of Kitchen Activities: the CMU-Multimodal Activity Database
  (CMU-MMAC)},'' \emph{CHIW}, 2009.

\bibitem{Trumble2017}
M.~Trumble, A.~Gilbert, C.~Malleson, A.~Hilton, and J.~Collomosse, ``{Total
  Capture: 3D Human Pose Estimation Fusing Video and Inertial Sensors},''
  \emph{BMVC}, 2017.

\bibitem{Chen2019}
T.~Chen, S.~Ding, J.~Xie, Y.~Yuan, W.~Chen, Y.~Yang, Z.~Ren, and Z.~Wang,
  ``{ABD-Net: Attentive but Diverse Person Re-Identification},'' \emph{ICCV},
  2019.

\bibitem{zhou2019}
K.~Zhou, Y.~Yang, A.~Cavallaro, and T.~Xiang, ``{Omni-Scale Feature Learning
  for Person Re-Identification},'' \emph{ICCV}, 2019.

\bibitem{Li2020}
Y.-J. Li, Z.~Luo, X.~Weng, and K.~M. Kitani, ``{Learning Shape Representations
  for Clothing Variations in Person Re-Identification},''
  \emph{arXiv:2003.07340}, 2020.

\bibitem{Li2019}
J.~Li, J.~Wang, Q.~Tian, W.~Gao, and S.~Zhang, ``{Global-Local Temporal
  Representations For Video Person Re-Identification},'' \emph{ICCV}, 2019.

\bibitem{Ren2015}
S.~Ren, K.~He, R.~Girshick, and J.~Sun, ``{Faster R-CNN: Towards Real-Time
  Object Detection with Region Proposal Networks},'' \emph{NIPS}, 2015.

\bibitem{Lee2016}
N.~Lee, X.~Weng, V.~N. Boddeti, Y.~Zhang, F.~Beainy, K.~Kitani, and T.~Kanade,
  ``{Visual Compiler: Synthesizing a Scene-Specific Pedestrian Detector and
  Pose Estimator},'' \emph{arXiv:1612.05234}, 2016.

\bibitem{Weng2018_r2n}
X.~Weng, S.~Wu, F.~Beainy, and K.~Kitani, ``{Rotational Rectification Network:
  Enabling Pedestrian Detection for Mobile Vision},'' \emph{WACV}, 2018.

\bibitem{Braun2019}
M.~Braun, S.~Krebs, F.~Flohr, and D.~M. Gavrila, ``{The EuroCity Persons
  Dataset: A Novel Benchmark for Object Detection},'' \emph{TPAMI}, 2019.

\bibitem{yolov3}
J.~Redmon and A.~Farhadi, ``{YOLOv3: An Incremental Improvement},''
  \emph{arXiv:1804.02767}, 2018.

\bibitem{xiu2018poseflow}
Y.~Xiu, J.~Li, H.~Wang, Y.~Fang, and C.~Lu, ``{{Pose Flow}: Efficient Online
  Pose Tracking},'' \emph{BMVC}, 2018.

\bibitem{hoffer2015}
E.~Hoffer and N.~Ailon, ``{Deep Metric Learning Using Triplet Network},''
  \emph{International Workshop on Similarity-Based Pattern Recognition}, 2015.

\bibitem{hermans2017}
A.~Hermans, L.~Beyer, and B.~Leibe, ``{In Defense of the Triplet Loss for
  Person Re-Identification},'' \emph{arXiv:1703.07737}, 2017.

\end{thebibliography}

\end{document}